\documentclass[conference]{IEEEtran}
\IEEEoverridecommandlockouts
\usepackage{cite}
\usepackage{graphicx}
\usepackage{amsmath}
\usepackage{amsfonts}
\usepackage{booktabs}
\usepackage{tikz}
\usetikzlibrary{arrows.meta}
\usepackage{balance}
\usepackage{url}
\usepackage{microtype}
\usepackage{xcolor}

\begin{document}
\title{COTTI: Contrastive Training and Time-Aware Inference for Robust Industrial Recommendation under Promotional Drift}
\author{
\IEEEauthorblockN{%
Maolei Huang\textsuperscript{1,*}\quad
Jing Tian\textsuperscript{2}\quad
Dexuan Duan\textsuperscript{2}\quad
Fangfang Lin\textsuperscript{3}\quad
Huawei Cao\textsuperscript{2}\quad
Shuhan Song\textsuperscript{2,*\textdagger}}
\IEEEauthorblockA{%
\textsuperscript{1}Alibaba Group, Beijing, China\\
\textsuperscript{2}University of Chinese Academy of Sciences, Beijing, China\\
\textsuperscript{3}Santa Clara University, USA\\[2pt]
\small\texttt{\{huangmaolei.hml, tianjing923.tj, ddx20010123, linfangfang\}@gmail.com}\\
\small\texttt{\{caohuawei, songshuhan\}@ict.ac.cn}}
\thanks{\textsuperscript{*}These authors contributed equally. \textsuperscript{\textdagger}Corresponding author: Shuhan Song (songshuhan@ict.ac.cn).}
}
\maketitle

\begin{abstract}
Promotional events shift traffic composition, purchase behavior, and the scale of conversion probabilities on different time horizons. We present \textbf{COTTI} (\textbf{CO}ntrastive \textbf{T}raining and \textbf{T}ime-aware \textbf{I}nference), an industrial recommendation framework that separates historical representation learning from event-time score adjustment. Cross-Period Contrastive Alignment (CPC) regularizes paired histories of the same user with adaptive consistency weights. Active Re-weighting (RW) controls the influence of promotional examples through historical-interest distance. Adaptive Boundary Alignment (ABA) combines stochastic prediction variability with a temporal event kernel to adjust logits while keeping predictor parameters fixed. We analyze the weighted objective and bound ABA-induced probability and ranking changes. On Fliggy data, COTTI attains an offline AUC of 0.8127, compared with 0.8101 for TEN and 0.8123 for TEN with ABA. Cross-category analysis reveals improved discrimination alongside declines in paid-item recall for novel categories. End-to-end measurements give a p99 latency of 38.9\,ms and throughput of 14,700 requests/s; enabling ABA adds 2.2\,ms to p99. A ten-day online experiment against the serving HDR model yields relative gains of 3.14\% in conversion rate and 2.68\% in revenue per mille. Together, these results characterize the trade-off between aggregate prediction, novel-category discovery, and serving cost in promotional recommendation.
\end{abstract}

\begin{IEEEkeywords}
E-commerce recommendation, adaptive recommendation, contrastive learning, distribution shift
\end{IEEEkeywords}

\section{Introduction}
\label{sec:intro}

Promotions change discounts, traffic sources, and purchase timing, altering the relationship between historical behavior and conversion. Models updated on routine traffic may respond too slowly to short promotional peaks. Event-aware adaptation aims to maintain ranking quality and reliable conversion-rate (CVR) estimates during these shifts.

Our design uses three sources of information. Regular-period history provides a reference for short-lived behavior without fully determining purchases during a sale. Promotional interactions include low-intent exploration and valuable discovery, both of which can depart from historical preferences. Announced campaign timing provides prospective context, although the direction and magnitude of individual responses remain uncertain.

Stability and adaptation can conflict: a user who usually books economy flights may purchase a premium hotel package during a promotion. Unconditional historical consistency could suppress this useful change, whereas excessive adaptation to transient interactions could weaken established preferences. We evaluate this trade-off on both the overall population and relevant subgroups.

Historical reuse and cross-temporal transfer address this setting from different directions. HDR selects and adapts historical promotion data \cite{chan2023capturing}; TEN aligns temporal representations and fuses promotion-sensitive experts \cite{wang2026temporalexpertnet}. COTTI coordinates adaptation at three levels: paired historical representations, individual promotional examples, and event-time conversion scores. This formulation connects historical regularization during training with uncertainty-conditioned score adjustment after predictor parameters are fixed.

The three components act on different objects. CPC encourages consistency between a user's historical views, but representation agreement alone does not control the influence of individual promotional examples. RW adjusts that influence while preserving conversion supervision, yet training-time weighting cannot respond to event timing after parameters are frozen. ABA supplies that serving-time adjustment using event context and predictive variability. This division motivates the combination without assuming additive effects. Our evaluation examines component removals, cross-category outcomes, simpler calibration controls, and end-to-end serving costs. It reveals a concrete trade-off: stronger discrimination can coexist with lower recall of purchased novel-category items.

Our contributions are as follows:
\begin{itemize}
  \item We formulate a three-component framework combining cross-period representation alignment, historical-distance example weighting, and event-conditioned score adjustment with fixed serving parameters.
  \item We characterize how exponential weighting changes the balance of the training objective and derive bounds on ABA-induced probability shifts and pairwise ranking changes.
  \item We evaluate component ablations, cross-category outcomes, calibration controls, serving costs, and an online comparison with HDR, identifying trade-offs between aggregate prediction, discovery, and deployment cost.
\end{itemize}

Robustness here refers to performance under the evaluated promotional conditions. The design imposes no distributionally robust or subgroup-level guarantee; its scope follows from the specified adaptation mechanisms and the comparisons actually evaluated.

\section{Related Work}

\subsubsection{Conversion Rate Prediction}
Industrial CVR models combine user, item, and context features to estimate purchase probability. Feature-interaction and multi-task architectures include Wide \& Deep, DeepFM, DCN, MMoE, and ESMM \cite{cheng2016wide,guo2017deepfm,wang2021dcn,ma2018modeling,ma2018entire}. Numerical-feature embeddings address another part of the prediction input \cite{guo2021embedding}. Attention and multi-interest modeling further represent heterogeneous user behavior \cite{zhou2018deep,zhou2019deep,pi2020search,li2019multi,cen2020controllable}. These modeling choices can improve prediction without, by themselves, specifying how to adapt to a short promotional event.

Delayed-feedback methods address the gap between an interaction and its eventual conversion label \cite{chapelle2014modeling,ktena2019addressing,yang2021capturing,chen2022asymptotically,huang2026dcrnet}. This issue is especially relevant when interpreting recent non-conversions: an interaction without an observed payment may still convert after the observation cutoff. Delayed feedback and promotional distribution shift are related but distinct, so label maturity must be controlled when comparing event-adaptation methods.

HDR couples historical-data retrieval, distribution correction, and fine-tuning \cite{chan2023capturing}. TEN uses BridgeNet for temporal representation alignment and TemporalExpertGate for sample-aware expert fusion, with a two-stage training strategy \cite{wang2026temporalexpertnet}. COTTI distinguishes the unit and stage of adaptation: same-user historical views for CPC, individual promotional examples for RW, and event-conditioned logits for ABA. This separation makes both training-time influence and serving-time score changes explicit.

\subsubsection{Contrastive Representation Learning}
Contrastive objectives encourage agreement between paired views while separating unpaired examples \cite{chen2020simple,he2020momentum,wang2020understanding}. In recommendation, the choice of views and positives determines which behavioral properties are encouraged to remain consistent; related work examines sequence similarity, cross-domain transfer, and contextual purchase behavior \cite{wang2024relative,chang2025separated,huang2025context}. Other industrial applications use hierarchical query categories and travel inventory-add signals \cite{zhu2023hcl4qc,li2025contrastive}. COTTI pairs a user's regular and promotional histories in an adaptively weighted InfoNCE objective, evaluated jointly with RW and ABA.

\subsubsection{Reweighting, Uncertainty, and Calibration}
Importance weighting under covariate shift relies on assumptions about how the training and test distributions differ \cite{sugiyama2007covariate}. Recommendation studies likewise distinguish exposure or observation bias from user preference \cite{schnabel2016treatments,saito2020unbiased}. COTTI's historical-distance weight is neither a density-ratio estimate nor an inverse exposure propensity. These works motivate explicit treatment of the prediction population and label-observation process, rather than an unbiasedness claim for RW.

MC dropout provides an approximate Bayesian interpretation of stochastic prediction \cite{gal2016dropout}; deep ensembles offer another route to predictive variability \cite{lakshminarayanan2017ensembles}. Neither construction automatically turns uncertainty into evidence of underprediction. Post-hoc calibration can improve probability estimates \cite{guo2017calibration}, but its reliability must also be checked under distribution shift \cite{ovadia2019uncertainty}. ABA therefore combines a specified uncertainty proxy with an event-conditioned directional correction whose value is assessed against simpler controls.

\section{Methodology}

We first define the prediction setting and then describe CPC, RW, and ABA. The presentation distinguishes quantities used for training from quantities computed during serving.

\subsection{Problem Definition}

Let $\mathcal U$ and $\mathcal I$ denote users and items. For user $u$, $\mathcal D_{u,r}$ indexes training targets in period $r\in\{\mathrm{reg},\mathrm{promo}\}$. Target $j$ has item $i_j$, time $t_j$, and label $y_{uj}$. Its features $x_{uj}$ use only history $S_u^{<t_j}$ and item/context information available at prediction time. The event calendar partitions that history into regular and promotional views. All representations and RW weights are computed for this pre-target cutoff; we suppress its index where unambiguous.

Promotional drift can change both the feature distribution and the relationship between features and conversion. As a descriptive measure, let $z_{u,t}=\mathcal A(f_\theta(S_u^{<t}))$ and define
\begin{equation}
\Delta(u)=\left\|\mathbb E_{t\in\mathcal T_{\mathrm{promo}}}[z_{u,t}]
-\mathbb E_{t\in\mathcal T_{\mathrm{reg}}}[z_{u,t}]\right\|_2^2.
\end{equation}
A large $\Delta(u)$ indicates a change in the mean representation, not necessarily a loss of ranking quality. The learning target remains conversion prediction over the promotional population:
\begin{equation}
\mathcal R_{\mathrm{promo}}(\theta,\phi)
=\mathbb E_{(x,y)\sim P_{\mathrm{promo}}}
  [\ell(y,p_{\theta,\phi}(x))],
\end{equation}
where $\ell$ is binary cross-entropy. We separate a learned base logit from an event-conditioned adjustment:
\begin{equation}
p_{\theta,\phi}(x)
=\sigma\!\left(h_\theta(x)+\delta_\phi(t,\mathbb U_\theta(x))\right).
\end{equation}
Here, $\theta$ denotes predictor parameters and $\phi$ the calibration settings. Historical consistency supplies a regularization prior; it does not define a latent noise label or replace the observed conversion target. In particular, a novel-category purchase remains a valid positive example even when it is distant from the historical profile.

Training and calibration selection are separate stages:
\begin{equation}
\begin{split}
\hat\theta&=\operatorname{Train}(\mathcal D;\theta_0),\\
\hat\phi&=\arg\min_{\phi\in\Phi}
   \mathcal L_{\mathrm{val}}(\hat\theta,\phi).
\end{split}
\label{eq:stages}
\end{equation}
Here, $\mathcal D$ is the training set and $\theta_0$ the initialization. The operator $\operatorname{Train}$ denotes mini-batch optimization with the gradient conventions in Section~\ref{sec:training}: RW weights are recomputed and detached at each update. The calibration settings $\phi=(\kappa,\beta)$ are then selected with the predictor fixed, over validation search space $\Phi$. Event-center and width information are obtained before the target event. This separation distinguishes learned predictor parameters from event-time score adjustment.

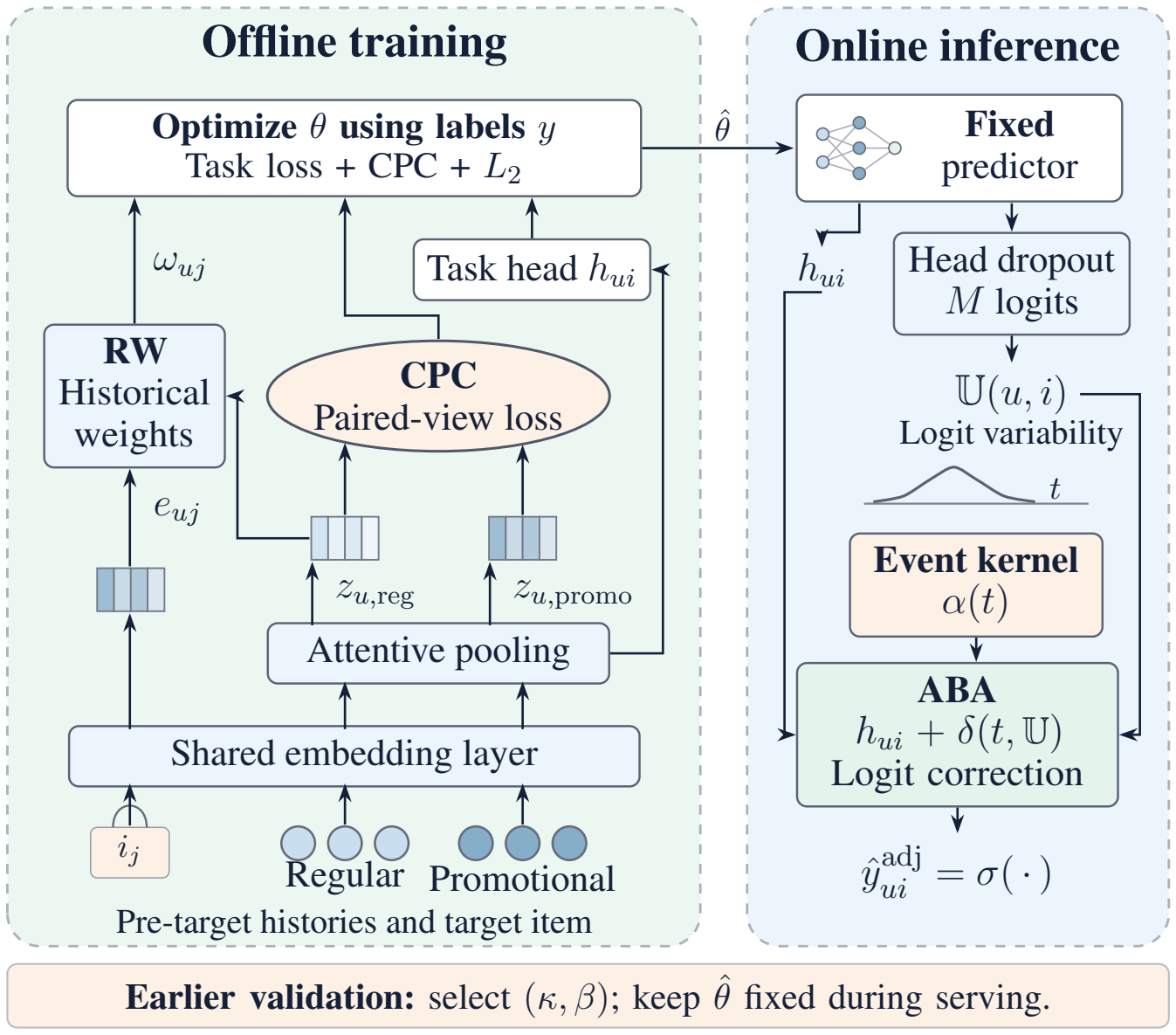
\begin{figure}[t]
  \centering
  \resizebox{\linewidth}{!}{%
  \begin{tikzpicture}[
    font=\rmfamily\fontsize{11.5}{12.6}\selectfont,
    text=cottiInk,
    flow/.style={-{Stealth[length=2mm,width=1.4mm]},draw=cottiInk,line width=.7pt},
    module/.style={draw=cottiInk!70,fill=white,rounded corners=3pt,
      align=center,line width=.65pt,inner sep=4pt},
    region/.style={rounded corners=10pt,dashed,line width=.75pt,draw=cottiInk!35}]
    \definecolor{cottiInk}{HTML}{142038}
    \definecolor{cottiBlue}{HTML}{ECF5FC}
    \definecolor{cottiGreen}{HTML}{E4F3EB}
    \definecolor{cottiPeach}{HTML}{FDF0E5}
    \definecolor{cottiPurple}{HTML}{7432AD}
    \definecolor{cottiReg}{HTML}{C9DFEF}
    \definecolor{cottiPromo}{HTML}{86AEC9}
    \path[use as bounding box] (0,0) rectangle (12.5,11.0);

    \filldraw[region,fill=cottiGreen!65] (.04,.92) rectangle (7.45,10.95);
    \filldraw[region,fill=cottiBlue] (7.95,.92) rectangle (12.46,10.95);
    \node[font=\rmfamily\bfseries\fontsize{14}{15}\selectfont] at (3.75,10.55) {Offline training};
    \node[font=\rmfamily\bfseries\fontsize{14}{15}\selectfont] at (10.2,10.55) {Online inference};

    \node[font=\rmfamily\fontsize{10.5}{11.5}\selectfont] at (3.75,1.15) {Pre-target histories and target item};
    \draw[fill=cottiPeach,draw=cottiInk!45,rounded corners=2pt] (.93,1.65) rectangle (1.77,2.19);
    \draw[draw=cottiInk!65,line width=.65pt] (1.17,2.19) .. controls (1.17,2.5) and (1.53,2.5) .. (1.53,2.19);
    \node[font=\rmfamily\fontsize{10}{11}\selectfont] at (1.35,1.94) {$i_j$};
    \foreach \x in {3.15,3.65,4.15} {
      \draw[fill=cottiReg,draw=cottiInk!65,line width=.7pt] (\x,2.02) circle (.18);
    }
    \foreach \x in {5.05,5.55,6.05} {
      \draw[fill=cottiPromo,draw=cottiInk!65,line width=.7pt] (\x,2.02) circle (.18);
    }
    \node at (3.65,1.65) {Regular};
    \node at (5.55,1.65) {Promotional};
    \node[module,fill=cottiBlue,minimum width=6.1cm,minimum height=.55cm] (emb) at (3.75,2.95) {Shared embedding layer};
    \draw[flow] (1.35,2.21) -- (1.35,2.66);
    \draw[flow] (3.65,2.22) -- (3.65,2.66);
    \draw[flow] (5.55,2.22) -- (5.55,2.66);

    \node[module,fill=cottiBlue,minimum width=3.65cm,minimum height=.55cm] (pool) at (4.65,4.05) {Attentive pooling};
    \draw[flow] (3.65,3.24) -- (3.65,3.75);
    \draw[flow] (5.55,3.24) -- (5.55,3.75);
    \draw[flow] (1.35,3.24) -- (1.35,4.48);

    \foreach \x/\shade in {0/80,1/45,2/65,3/30} {
      \filldraw[fill=cottiPromo!\shade,draw=cottiInk!60,line width=.5pt]
        (1.00+\x*.18,4.50) rectangle (1.18+\x*.18,4.96);
      \filldraw[fill=cottiReg!\shade,draw=cottiInk!60,line width=.5pt]
        (3.29+\x*.18,5.05) rectangle (3.47+\x*.18,5.51);
      \filldraw[fill=cottiPromo!\shade,draw=cottiInk!60,line width=.5pt]
        (5.19+\x*.18,5.05) rectangle (5.37+\x*.18,5.51);
    }
    \node[anchor=west] at (1.48,5.56) {$e_{uj}$};
    \node[anchor=west] at (3.42,4.68) {$z_{u,\mathrm{reg}}$};
    \node[anchor=west] at (5.32,4.68) {$z_{u,\mathrm{promo}}$};
    \draw[flow] (3.30,4.35) -- (3.30,5.03);
    \draw[flow] (5.20,4.35) -- (5.20,5.03);

    \draw[fill=cottiPeach,draw=cottiInk!75,line width=.8pt] (4.65,6.8) ellipse (1.83 and .59);
    \node[align=center] (cpc) at (4.65,6.8) {\textbf{CPC}\\Paired-view loss};
    \node[module,fill=cottiBlue,text width=1.65cm,minimum height=1.02cm] (rw) at (1.4,6.8) {\textbf{RW}\\Historical\\weights};
    \draw[flow] (3.65,5.53) -- (3.65,6.30);
    \draw[flow] (5.55,5.53) -- (5.55,6.28);
    \draw[flow] (1.35,4.98) -- (1.35,4.98 |- rw.south);
    \draw[flow] (3.27,5.28) -- (2.5,5.28) |- (rw.east);

    \node[module,minimum width=2.12cm,minimum height=.65cm] (head) at (5.65,8.15) {Task head $h_{ui}$};
    \draw[flow] (pool.east) -- (7.05,4.05) |- (head.east);
    \node[module,fill=white,text width=5.85cm,minimum height=.86cm,font=\rmfamily\fontsize{11}{12}\selectfont] (loss) at (3.75,9.45)
      {\textbf{Optimize $\theta$ using labels $y$}\\Task loss + CPC + $L_2$};
    \draw[flow] (rw.north) -- (rw.north |- loss.south);
    \node[anchor=west] at (1.48,8.18) {$\omega_{uj}$};
    \draw[flow] (4.65,7.41) -- (4.65,7.66) -- (3.65,7.66) -- (3.65,7.66 |- loss.south);
    \draw[flow] (head.north) -- (head.north |- loss.south);

    \draw[flow] (loss.east) -- (8.45,9.45);
    \node[fill=white,inner sep=1pt,font=\rmfamily\fontsize{10}{11}\selectfont] at (7.69,9.7) {$\hat\theta$};
    \node[module,fill=white,minimum width=3.48cm,minimum height=1.14cm] (pred) at (10.22,9.45) {};
    \foreach \ya in {9.3,9.6} {
      \foreach \yb in {9.18,9.45,9.72} {
        \draw[draw=cottiInk!35,line width=.45pt] (8.76,\ya) -- (9.15,\yb);
      }
    }
    \foreach \ya in {9.18,9.45,9.72} {
      \draw[draw=cottiInk!35,line width=.45pt] (9.15,\ya) -- (9.53,9.45);
      \filldraw[fill=cottiPromo,draw=cottiInk!65,line width=.45pt] (9.15,\ya) circle (.065);
    }
    \foreach \ya in {9.3,9.6} {
      \filldraw[fill=cottiReg,draw=cottiInk!65,line width=.45pt] (8.76,\ya) circle (.065);
    }
    \filldraw[fill=cottiGreen,draw=cottiInk!65,line width=.45pt] (9.53,9.45) circle (.065);
    \node[align=center] at (10.76,9.45) {\textbf{Fixed}\\predictor};

    \node at (8.75,8.15) {$h_{ui}$};
    \node[module,fill=cottiBlue,minimum width=2.52cm,minimum height=.92cm] (mc) at (10.78,8.0)
      {Head dropout\\$M$ logits};
    \draw[flow] (9.15,8.86) -- (9.15,8.55) -- (8.75,8.55) -- (8.75,8.39);
    \draw[flow] (10.78,8.86) -- (mc.north);
    \node[font=\rmfamily\fontsize{13}{14}\selectfont] (unc) at (10.78,6.82) {$\mathbb U(u,i)$};
    \draw[flow] (mc.south) -- (unc.north);
    \node[font=\rmfamily\fontsize{10.5}{11.5}\selectfont] at (10.78,6.38) {Logit variability};

    \node[module,fill=cottiPeach,minimum width=2.7cm,minimum height=.93cm] (kernel) at (10.4,4.78)
      {\textbf{Event kernel}\\$\alpha(t)$};
    \draw[draw=cottiInk!55,line width=.6pt] (9.2,5.65) -- (11.6,5.65);
    \draw[draw=cottiInk!70,line width=.85pt,smooth]
      plot coordinates {(9.3,5.67) (9.62,5.73) (9.91,5.91) (10.16,6.04) (10.4,5.91) (10.7,5.73) (11.03,5.67)};
    \node[anchor=west,font=\rmfamily\fontsize{10}{11}\selectfont] at (11.06,5.81) {$t$};

    \node[module,fill=cottiGreen,text width=3.15cm,minimum height=1.25cm] (aba) at (10.2,3.18)
      {\textbf{ABA}\\$h_{ui}+\delta(t,\mathbb U)$\\Logit correction};
    \draw[flow] (8.75,7.91) -- (8.35,7.91) |- (aba.west);
    \draw[flow] (unc.east) -- (12.15,6.82) -- (12.15,3.18) -- (aba.east);
    \draw[flow] (kernel.south) -- (kernel.south |- aba.north);
    \node[font=\rmfamily\fontsize{13}{14}\selectfont] (score) at (10.2,1.72) {$\hat y^{\mathrm{adj}}_{ui}=\sigma(\,\cdot\,)$};
    \draw[flow] (aba.south) -- (score.north);

    \filldraw[fill=cottiPeach,draw=cottiInk!30,rounded corners=3pt,line width=.55pt]
      (.04,.05) rectangle (12.46,.73);
    \node[font=\rmfamily\fontsize{11}{12}\selectfont] at (6.25,.39)
      {\textbf{Earlier validation:} select $(\kappa,\beta)$; keep $\hat\theta$ fixed during serving.};
  \end{tikzpicture}}
  \caption{COTTI workflow. CPC and RW shape the predictor; earlier validation selects calibration settings; ABA adjusts scores with fixed predictor parameters during serving.}
  \label{fig:cranet}
  \end{figure}

\subsection{Representation Alignment via Cross-Period Contrast}
\label{CPC}

CPC encourages regular and promotional histories of the same user to retain shared information. This is an inductive preference for continuity, not an assumption that promotional interests must equal historical interests.

\subsubsection{Dual-View Representation Learning}
A shared encoder $f_\theta$ and attentive pooling layer $\mathcal A$ produce
\begin{equation}
z_{u,\mathrm{reg}}=\mathcal A(f_\theta(S_{u,\mathrm{reg}})),\quad
z_{u,\mathrm{promo}}=\mathcal A(f_\theta(S_{u,\mathrm{promo}})).
\end{equation}
Shared parameters make the views comparable. Let $\mathcal U_{\mathrm{pair}}$ denote users with both pre-target histories available for contrastive pairing. This eligibility condition applies to CPC, rather than defining the entire supervised or serving population.

The task head maps user, item, and context features to a conversion logit. The historical view separately supplies the training-time RW reference.

\subsubsection{Cross-Period Contrastive Objective}
The two views of user $u$ form a positive pair. Promotional views of other users provide negative examples:
{\small
\begin{equation}
\ell_{\mathrm{cpc}}(u)=
-\log\frac{\exp(\mathrm{s}(z_{u,\mathrm{reg}},z_{u,\mathrm{promo}})/\tau)}
{\sum_{v\in\mathcal B_u}\exp(\mathrm{s}(z_{u,\mathrm{reg}},z_{v,\mathrm{promo}})/\tau)}.
\end{equation}
}
Here, $\mathcal B_u\subseteq\mathcal U_{\mathrm{pair}}$ contains $u$ and sampled negative users, $\mathrm{s}$ is cosine similarity, and $\tau$ is the temperature. Since different users may share interests, the consistency loss must be balanced against task supervision.

\subsubsection{Adaptive Consistency Weight}
To modulate alignment strength across users, CPC uses
{\small
\begin{equation}
\begin{split}
\mathcal L_{\mathrm{cpc}}^{\mathrm{weighted}}
&=\sum_{u\in\mathcal U_{\mathrm{pair}}}\eta_u\ell_{\mathrm{cpc}}(u),\\
\eta_u&=1-\sigma(v_u).
\end{split}
\end{equation}
}
For representations of dimension $d_z$, let $\bar z_u=(z_{u,\mathrm{reg}}+z_{u,\mathrm{promo}})/2$. We compute the population variance of the two views in each coordinate and average across coordinates:
\begin{equation}
\begin{split}
v_u&=\frac{1}{2d_z}\sum_{r\in\{\mathrm{reg},\mathrm{promo}\}}
\|z_{u,r}-\bar z_u\|_2^2\\
&=\frac{\|z_{u,\mathrm{reg}}-z_{u,\mathrm{promo}}\|_2^2}{4d_z}.
\end{split}
\end{equation}
Thus, $\eta_u\in(0,0.5]$ decreases with the squared distance between views. Unlike RW weights, $\eta_u$ participates in backpropagation: $\nabla_\theta(\eta_u\ell_{\mathrm{cpc}})=\eta_u\nabla_\theta\ell_{\mathrm{cpc}}+\ell_{\mathrm{cpc}}\nabla_\theta\eta_u$. Minimization can therefore improve agreement or reduce the gate, rather than enforcing uniform alignment. Representation scale affects the gate, and distance does not identify which departures correspond to useful discovery. Section~\ref{sec:subgroup} examines the resulting trade-off.

\subsection{Historical-Consistency Reweighting}
\label{RW}

RW changes the influence of promotional examples according to their distance from historical preferences. Its motivation is to reduce sensitivity to highly dispersed interactions when they are weak predictors of conversion. Distance alone cannot distinguish low-intent browsing from valuable cross-category discovery; the method therefore retains conversion labels as the supervision signal.

\subsubsection{Historical Divergence}
For promotional target $j$ of user $u$, let $e_{uj}$ denote the item representation in the space used by the pre-target historical user representation. RW computes
\begin{equation}
d_{uj}=\|e_{uj}-z_{u,\mathrm{reg}}\|_2^2.
\end{equation}
RW requires compatible item and historical-user representations. Their projections and normalization determine the meaning and scale of distance; distance alone does not determine whether a purchase is valuable.

\subsubsection{Relative Influence Weights}
The exponential weighting function is
\begin{equation}
\omega_{uj}=\exp\left(-\gamma_{\mathrm{rw}}\,
\frac{d_{uj}-\mu_b}{\sigma_b+\epsilon}\right),
\label{gamma}
\end{equation}
where $\mu_b$ and $\sigma_b$ are the mini-batch mean and standard deviation of divergence, and $\epsilon$ stabilizes the denominator. We use these weights directly, without clipping or mean normalization. Below-average distances receive weights above one and above-average distances receive weights below one. These positive influence weights are not reliability probabilities.

\subsubsection{Weighted Supervision}
The promotional task loss is
\begin{equation}
\mathcal L_{\mathrm{rw}}=\sum_{u\in\mathcal U}
\sum_{j\in\mathcal D_{u,\mathrm{promo}}}\operatorname{sg}(\omega_{uj})\ell(y_{uj},\hat y_{uj}),
\end{equation}
with binary cross-entropy $\ell$. The weight does not depend explicitly on $y_{uj}$, so paid interactions can also be downweighted. Positive weights preserve a nonzero contribution, but not necessarily sufficient influence for novel purchases.

The stop-gradient operator $\operatorname{sg}$ preserves its argument in the forward computation and has zero derivative. A promotional sample therefore contributes $\omega_{uj}\nabla_\theta\ell$, without the additional term $\ell\nabla_\theta\omega_{uj}$. Weights are recomputed from the current representations at each update. This convention retains distance-dependent task influence while preventing direct optimization through the RW weight.

\subsection{Adaptive Boundary Alignment}
\label{ABA}

ABA applies a bounded, event-conditioned adjustment to conversion logits. The predictor parameters remain fixed; the adjustment changes scores through the current event context and stochastic prediction variability.

\subsubsection{Predictive Variability}
For pair $(u,i)$, let $h_{ui}^{(m)}$ be the pre-sigmoid logit under dropout mask $m$, and let $\bar h_{ui}$ be the mean of $M$ samples. We use
\begin{equation}
\mathbb U(u,i)=\frac{1}{M}\sum_{m=1}^{M}
\left(h_{ui}^{(m)}-\bar h_{ui}\right)^2
\end{equation}
as a proxy for predictive uncertainty. All samples share a deterministic encoder output; dropout is applied only in the prediction head. The variance therefore measures sensitivity to head perturbations, not encoder uncertainty, distribution shift, or purchase intent. Section~\ref{sec:deployment} evaluates its serving cost.

\subsubsection{Temporal Event Context}
A Gaussian kernel describes the temporal intensity of the correction:
\begin{equation}
\alpha(t)=\kappa\exp\left(-\frac{(t-\mathcal T_{\mathrm{peak}})^2}{2s_{\mathrm{event}}^2}\right).
\end{equation}
The event center $\mathcal T_{\mathrm{peak}}$ is fixed by the announced campaign schedule. The width $s_{\mathrm{event}}$ is determined from high-traffic periods of historical campaigns, using their top 20\% of peak-traffic hours as the reference interval. Both are available before the target event; the current test period's realized traffic peak is not used. The non-negative amplitude $\kappa$ is selected by validation grid search. The exact mapping from the historical interval to Gaussian width, the amplitude grid, and the selected amplitude are not specified here, which limits exact reproduction of the temporal correction.

\subsubsection{Event-Conditioned Logit Adjustment}
Let $h_{ui}$ be the deterministic prediction logit computed with dropout disabled. ABA returns
\begin{equation}
\hat y_{ui}^{\mathrm{adj}}=
\sigma\!\left(h_{ui}+\alpha(t)\tanh(\beta\mathbb U(u,i))\right),
\end{equation}
where $\beta>0$. For $\kappa\ge0$, the correction can only increase the predicted probability. The one-sided correction targets underprediction and cannot reduce overestimated probabilities. Predictive variability controls the correction's magnitude but does not justify its direction.

A time-only bias corrects an event-level offset, whereas uncertainty can differentiate pairs. Section~\ref{sec:calibration} tests the value of this variation. Since aggregate underprediction can coexist with subgroup overprediction, ranking gains and calibration require separate assessment.

\subsection{Training Objective and Calibration Selection}
\label{sec:training}

Using probabilities $\hat y_{uj}$ before ABA, task supervision combines regular and reweighted promotional targets:
\begin{equation}
\begin{split}
\mathcal L_{\mathrm{task}}(u)
={}&\sum_{j\in\mathcal D_{u,\mathrm{reg}}}\ell(y_{uj},\hat y_{uj})\\
&+\sum_{j\in\mathcal D_{u,\mathrm{promo}}}
\operatorname{sg}(\omega_{uj})\ell(y_{uj},\hat y_{uj}).
\end{split}
\end{equation}
The per-update training objective is
\begin{equation}
\mathcal L_{\mathrm{train}}=
\sum_{u\in\mathcal U}\mathcal L_{\mathrm{task}}(u)
+\lambda_{\mathrm{cpc}}\mathcal L_{\mathrm{cpc}}^{\mathrm{weighted}}
+\lambda_2\|\theta\|^2.
\end{equation}
The coefficients $\lambda_{\mathrm{cpc}}$ and $\lambda_2$ control contrastive regularization and parameter shrinkage, separately from the RW scale $\gamma_{\mathrm{rw}}$. Each update treats the current RW weights as constants while differentiating through CPC and its gate. This defines an iteratively reweighted optimization procedure, rather than unconstrained differentiation of every forward dependency. Calibration follows training with the predictor fixed, as in Eq.~\ref{eq:stages}.

\subsubsection{Implementation Scope}
The specification does not fully define history construction, missing-view fallbacks, task-head fusion, the negative pool, or selected loss coefficients. In particular, CPC eligibility alone does not define RW behavior for users without a regular-period reference; that fallback is necessary for full-population reproduction.

\subsection{Properties of Weighting and Score Adjustment}
\label{sec:properties}

\subsubsection{Conditional Risk under Fixed Weights}
Let $x$ contain the prediction features and $p(x)=P(Y=1\mid x)$. In an idealized setting with a fixed, positive, label-independent weight $w(x)$, the conditional weighted cross-entropy is
\begin{equation}
\begin{split}
r_x(q)&=-w(x)\{p(x)\log q+[1-p(x)]\log(1-q)\},\\
\frac{\partial r_x}{\partial q}&=w(x)\frac{q-p(x)}{q(1-q)}.
\end{split}
\label{eq:conditional_weight}
\end{equation}
For $0<p(x)<1$, the minimizer remains $q=p(x)$ if predictions can vary freely at each $x$. COTTI need not attain this ideal: shared parameters, CPC, and batch-dependent weights couple optimization across examples even with detached weights. Its subgroup behavior must be measured.

Within a batch, two RW weights satisfy
\begin{equation}
\frac{\omega_a}{\omega_b}
=\exp\left[-\gamma_{\mathrm{rw}}\frac{d_a-d_b}{\sigma_b+\epsilon}\right].
\label{eq:weight_ratio}
\end{equation}
With the RW scale 1.5, a difference of two normalized distance units produces a weight ratio of $e^3\approx20.1$ in favor of the nearer example. This algebraic example illustrates why weight distributions and paid-example influence matter even when the aggregate ablation is positive.

\subsubsection{Weight Scale and Loss Balance}
Batch standardization does not normalize the mean influence weight. Let $a_j=(d_j-\mu_b)/(\sigma_b+\epsilon)$. Since the batch mean of $a_j$ is zero, convexity of the exponential gives $\bar\omega_b=n_b^{-1}\sum_j\exp(-\gamma_{\mathrm{rw}}a_j)\ge1$. The inequality is strict when the normalized distances vary and the scale is positive. Thus, the unmodified exponential rule changes both the relative influence of examples and the overall scale of promotional supervision.

Because the implementation does not normalize weights, increasing their mean also increases promotional supervision relative to regular-period loss, CPC, and parameter regularization. Dividing by $\bar\omega_b$ would preserve within-batch ratios while changing that balance; this diagnostic alternative has not been evaluated.

\subsubsection{Bounded Score Changes and Ranking}
Let $\delta_{ui}(t)$ denote the additive ABA logit correction. It equals $\alpha(t)\tanh(\beta\mathbb U(u,i))$. For non-negative $\kappa$ and $\mathbb U$,
\begin{equation}
\begin{split}
0&\le\delta_{ui}(t)\le\alpha(t)\le\kappa,\\
0&\le\hat y_{ui}^{\mathrm{adj}}-\sigma(h_{ui})
\le\min\{1,\alpha(t)/4\}.
\end{split}
\label{eq:shift_bound}
\end{equation}
The probability bound follows from $\sigma'(h)=\sigma(h)[1-\sigma(h)]\le1/4$. The corresponding odds are multiplied by $\exp(\delta_{ui})$. These bounds limit the size of an adjustment, not its calibration error or business effect.

A common time-only bias preserves ordering within a request. ABA changes ordering through candidate-specific uncertainty, but preserves a pair if $h_i-h_j>\alpha(t)$ because $\delta_j-\delta_i\le\alpha(t)$. Reordering is thus confined to small logit margins. This bound does not establish that reordering improves ranking or calibration. At cutoff $q_0\in(0,1)$, acceptance becomes $h_{ui}\ge\operatorname{logit}(q_0)-\delta_{ui}(t)$; no threshold is learned within the request.

\subsubsection{Execution Order and Applicability}
Construct each training target's histories before its prediction time, then optimize task, RW, and CPC terms. Freeze the predictor before selecting calibration on earlier validation data. Serving computes logits and stochastic variability through the actual dropout path, then applies the temporal correction. CPC pairs and training-batch RW statistics are unnecessary within a request.

The design is most plausible when historical interests contain reusable information and announced event timing predicts systematic underprediction. It is less suitable when stable histories are absent, behavior shifts permanently, or the base model overpredicts. Algebraically, $\kappa=0$ or $\mathbb U=0$ recovers the deterministic base probability $\sigma(h_{ui})$, and the Gaussian correction vanishes far from the event center. These limits describe the score function; an automatic fallback or drift-detection policy is outside the evaluated design.

\section{Experiments}

Our evaluation addresses three questions: whether COTTI improves aggregate prediction, how historical-consistency regularization affects useful discovery, and what predictive gains and serving costs ABA adds beyond simpler calibration. We examine these questions through offline comparisons, component ablations, calibration controls, online evaluation, and serving measurements.

\subsection{Experimental Setup}

\subsubsection{Data and Prediction Target}
The dataset comprises multi-month impression and conversion logs from Fliggy, covering regular traffic and the 6.18, Double 11, and Double 12 promotional periods. Table~\ref{tab:data} summarizes its approximate scale across traffic regimes. Population scope, deduplication, and averaging intervals are not fully specified. The prediction denominator, payment attribution, and label-maturation window are also incompletely specified. This limits reproducibility of the target population and negative labels, since an unobserved conversion can represent delayed feedback rather than a mature negative.

\begin{table}[t]
\centering
\caption{Approximate data scale across regular, preheating, and promotion-day traffic. Aggregation scope is discussed in the experimental setup.}
\label{tab:data}
\begin{tabular}{lccc}
\toprule
Scenario & \#Users & \#Items & \#Samples\\
\midrule
Daily (avg.) & $\approx$10M & $\approx$100K & $\approx$50M\\
Preheating (avg.) & $\approx$15M & $\approx$200K & $\approx$80M\\
Promotion day (avg.) & $\approx$20M & $\approx$300K & $\approx$100M\\
\bottomrule
\end{tabular}
\end{table}

\subsubsection{Chronological Protocol}
The offline evaluation uses over 100 million samples from November 11, 2025. December 1--10, 2025 is the online A/B observation window, followed by full deployment on December 12. The December observations are excluded from tuning the November offline model and its calibration parameters. Table~\ref{tab:timeline} separates these evaluation stages. Exact earlier training and validation cutoffs are not specified.

\begin{table}[t]
\centering
\caption{Evaluation chronology. Offline prediction, online A/B testing, and full deployment are distinct stages.}
\label{tab:timeline}
\small
\setlength{\tabcolsep}{3pt}
\begin{tabular}{lll}
\toprule
Stage & Evaluation & Evidence\\
\midrule
Offline & Nov.\ 11, 2025 & Prediction quality\\
Online A/B & Dec.\ 1--10, 2025 & HDR comparison\\
Full deployment & Dec.\ 12, 2025 & Full-traffic trace\\
\bottomrule
\end{tabular}
\end{table}

The pre-target feature construction in Section~\ref{CPC} excludes future interactions from COTTI's histories. Historical-data access and target-event adaptation also affect the comparators. Their method-specific cutoffs and adaptation budgets are not fully specified, so the system-level contrasts do not isolate architecture from all data-access and optimization choices.

\subsubsection{Metrics and Statistical Reporting}
AUC measures discrimination between converted and non-converted examples \cite{bradley1997use}; cross-entropy loss and RMSE assess probability errors. PCOC is the ratio of aggregate predicted conversions to observed conversions, with one as its target. PCOC can conceal offsetting errors across users or periods. Our probability-quality evaluation covers aggregate error and PCOC; it does not provide a probability-bin or event-time reliability analysis.

Online CVR and revenue per mille (RPM) measure purchase and revenue outcomes; RPM is attributed revenue per thousand recommendation impressions. The CVR denominator, refund treatment, and attribution windows are not fully specified. Online lifts are relative; offline AUC differences are absolute. We report point estimates without confidence intervals, repeated-run variability, or significance claims. Large impression counts do not establish precision when observations are correlated within users or periods.

\subsubsection{Implementation}
Our implementation uses TensorFlow, a shared four-layer MLP with hidden dimension 512, Adam with learning rate $10^{-4}$, 5,000 warmup iterations, and batch size 2048. The CPC temperature is $\tau=0.07$ and the RW scale is $\gamma_{\mathrm{rw}}=1.5$. The ABA sensitivity grid is $\beta\in\{0.1,0.5,1.0,2.0\}$; stochastic prediction uses $M=10$ and dropout rate 0.2. Selected loss and calibration coefficients, the stopping rule, and predictor-training search budgets are not fully specified. The matched budgets in Section~\ref{sec:calibration} concern post-training adjustment, not predictor training. The training environment includes 64 NVIDIA A100 GPUs; Section~\ref{sec:deployment} discusses resource accounting.

\subsubsection{Comparators}
We compare COTTI with production, promotion-transfer, and delayed-feedback methods, together with component removals:
\begin{itemize}
\item \textbf{Base} is the production ESMM-based predictor \cite{ma2018entire}, incrementally updated on daily traffic. \textbf{Promotion transfer} applies a model trained on a historical promotion to a target period; \textbf{promotion indicator} adds event-context features.
\item \textbf{HDR} retrieves historical promotional examples, corrects their distribution through reweighting, and adapts the predictor with its TransBlock component \cite{chan2023capturing}. Its historical-data access is part of the comparison protocol.
\item \textbf{ES-DFM} and \textbf{FNW} address delayed-feedback bias through sampling or reweighting \cite{yang2021capturing,ktena2019addressing}. These approaches target label timing rather than event-conditioned score adjustment.
\item \textbf{TEN} combines cross-temporal representation alignment with promotion-sensitive expert fusion \cite{wang2026temporalexpertnet}. It is the strongest non-COTTI comparator by AUC in Table~\ref{tab:model_variants}; Table~\ref{tab:calibration} additionally evaluates TEN with ABA.
\item \textbf{COTTI w/o CPC}, \textbf{w/o RW}, and \textbf{w/o ABA} remove cross-period alignment, replace promotional weights with equal weights, or disable inference-stage score adjustment, respectively.
\end{itemize}
Baseline code versions and complete adaptation schedules are not specified. We therefore interpret these comparisons as evaluations of the implemented systems under the described protocol, rather than isolated tests of architecture with every training choice controlled.

\subsection{Offline Results and Component Ablations}
\label{sec:offline}

\begin{table}[t]
\centering
\small
\caption{Offline prediction results. Bold entries mark the best ranking and probability-error values; PCOC is interpreted by distance from one.}
\label{tab:model_variants}
\setlength{\tabcolsep}{5.5pt}
\begin{tabular}{lcccc}
\toprule
Model & AUC & RMSE & LOSS & PCOC\\
\midrule
Base & 0.7902 & 0.1901 & 0.1591 & 0.9293\\
\quad w/ promotion transfer & 0.7993 & 0.1893 & 0.1510 & 1.0640\\
\quad w/ promotion indicator & 0.7981 & 0.1896 & 0.1503 & 1.0510\\
\midrule
HDR & 0.8001 & 0.1884 & 0.1509 & 0.9306\\
ES-DFM & 0.8010 & 0.1888 & 0.1513 & 0.9311\\
FNW & 0.8014 & 0.1828 & 0.1512 & 0.9316\\
TEN & 0.8101 & 0.1849 & 0.1503 & 0.9304\\
\midrule
\textbf{COTTI (Full)} & \textbf{0.8127} & \textbf{0.1807} & \textbf{0.1482} & 0.9432\\
\quad w/o CPC & 0.8082 & 0.1821 & 0.1488 & 0.9336\\
\quad w/o RW & 0.8051 & 0.1809 & 0.1489 & 0.9328\\
\quad w/o ABA & 0.8102 & 0.1830 & 0.1496 & 0.9366\\
\bottomrule
\end{tabular}
\end{table}

COTTI attains the highest AUC and lowest RMSE and cross-entropy in Table~\ref{tab:model_variants}. Relative to TEN without ABA, the AUC difference is 0.0026, or 0.26 percentage points. The stronger TEN-with-ABA control narrows that difference to 0.0004 (Section~\ref{sec:calibration}), distinguishing improvement over the unadjusted comparator from improvement when both systems use ABA.

Discrimination and aggregate calibration favor different models. COTTI's PCOC of 0.9432 indicates residual underprediction. The promotion-indicator model has a smaller absolute PCOC error, 0.0510 compared with 0.0568 for COTTI, despite its lower AUC. The ranking gains therefore coexist with a less accurate aggregate conversion total than this comparator.

The component removals characterize effects within the combined configuration. Removing CPC reduces AUC from 0.8127 to 0.8082. Removing RW reduces it to 0.8051, while cross-entropy rises from 0.1482 to 0.1489. Removing ABA gives AUC 0.8102 and PCOC 0.9366. These contrasts support the aggregate usefulness of the components, without separating their interactions. Sections~\ref{sec:subgroup} and~\ref{sec:calibration} examine whether the gains extend to discovery outcomes and simpler score-adjustment controls.

\begin{figure*}[t]
\centering
\includegraphics[width=\linewidth]{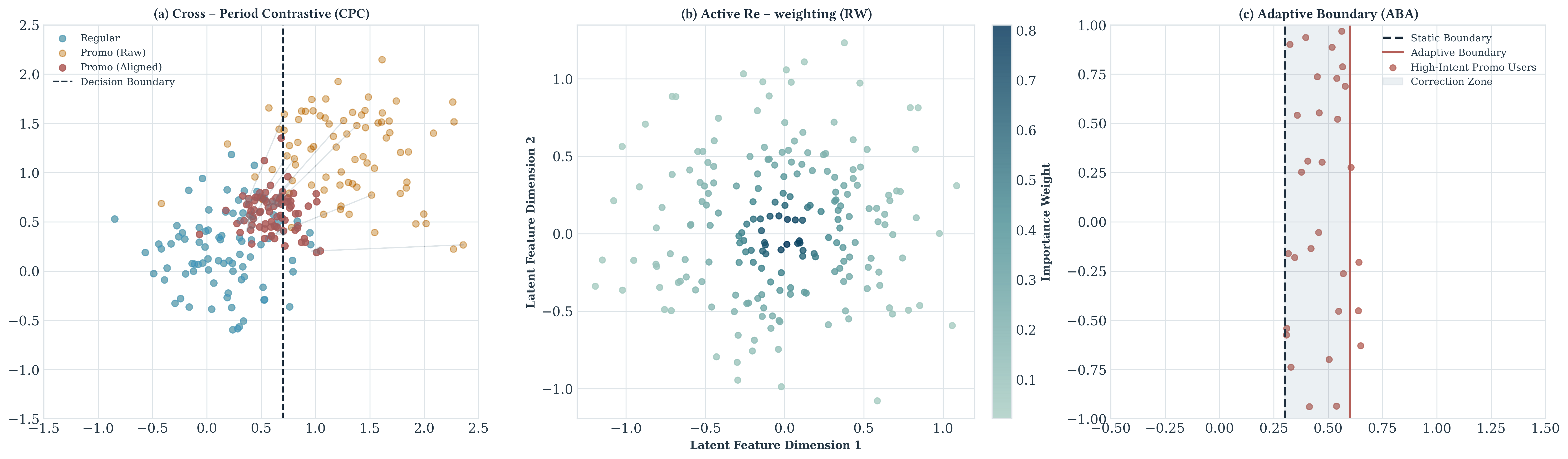}
\caption{Schematic illustration of the three mechanisms: CPC encourages cross-period consistency, RW adjusts influence by historical distance, and ABA modifies event-time scores.}
\label{fig:2}
\end{figure*}

\subsection{Cross-Category Conversion Analysis}
\label{sec:subgroup}

The cross-category analysis examines whether aggregate RW gains coexist with reduced support for useful discovery. A candidate is novel-category for a user when its category is absent from that user's pre-impression history; otherwise it is familiar-category. The category level, historical lookback, cohort and conversion counts, and label-maturation window are not specified. The contrasts therefore describe the evaluated cohorts without establishing how broadly those cohorts cover the serving population.

For cohorts containing both labels, AUC measures ranking discrimination. Positive-class log loss, $\mathrm{LL}^{+}=-|\mathcal P|^{-1}\sum_{j\in\mathcal P}\log\hat y_j$, measures probability support for the purchased items in set $\mathcal P$. Paid-item recall measures whether an observed purchase appears among the top $K$ candidates. The recall cutoff and candidate-set construction are not specified, limiting independent reproduction of these contrasts. Recall from logged candidates is an offline ranking measure, not a causal estimate of incremental purchases.

\begin{table}[t]
\centering
\small
\caption{Measured cross-category differences (Full minus the named ablation). Higher AUC/recall and lower positive-class loss are preferable. All differences are absolute.}
\label{tab:subgroup}
\setlength{\tabcolsep}{3pt}
\begin{tabular}{lccc}
\toprule
Cohort / reference & $\Delta$AUC & $\Delta$LL$^+$ & $\Delta$Recall\\
\midrule
Familiar / w/o RW & +0.0082 & -0.0120 & +0.0034\\
Novel / w/o RW & +0.0028 & +0.0038 & -0.0008\\
Novel / w/o CPC & +0.0016 & +0.0019 & -0.0004\\
\bottomrule
\end{tabular}
\end{table}

Table~\ref{tab:subgroup} shows that RW improves all three reported metrics for familiar categories: AUC rises by 0.0082, positive-class loss falls by 0.0120, and paid-item recall rises by 0.0034. For novel categories, the full model improves AUC by 0.0028 over w/o RW but increases positive-class loss by 0.0038 and reduces recall by 0.0008. The comparison with w/o CPC follows the same pattern: AUC improves by 0.0016, while loss rises by 0.0019 and recall falls by 0.0004. The recall reductions are 0.08 and 0.04 percentage points, respectively, rather than relative percentage changes.

AUC measures positive--negative ordering across a cohort, whereas positive-class loss and paid-item recall focus on observed purchases and the cutoff. Better discrimination can therefore coexist with worse purchase support or recall. These results support aggregate usefulness, not protection of novel-category discovery; without cohort counts and paired intervals, the precision of the differences remains unknown.

Weight quantiles and the share of total RW weight assigned to paid novel-category examples would help distinguish reduced training influence from other causes of this pattern. A joint CPC/RW removal on fixed cohorts would additionally test interactions between alignment and weighting. Neither diagnostic is available, so the subgroup contrasts should not be interpreted as identifying the causal mechanism of the recall decrease.

\subsection{Calibration Controls}
\label{sec:calibration}

The proximity of COTTI w/o ABA (0.8102 AUC) to TEN (0.8101) motivates a direct examination of score adjustment. Table~\ref{tab:calibration} compares simpler alternatives and applies ABA to TEN. A time-only alternative uses $h_{ui}+b(t)$ with event-conditioned bias but no uncertainty term. Affine calibration uses a validation-fitted scale and intercept; the uncertainty-only alternative replaces the event kernel with a constant amplitude. All controls use the same offline test set, validation interval, and tuning budget, with predictor parameters fixed during calibration.

\begin{table}[t]
\centering
\small
\caption{Measured calibration controls with fixed predictors and matched validation and tuning budgets. Bold entries mark the best values.}
\label{tab:calibration}
\setlength{\tabcolsep}{3pt}
\begin{tabular}{llcc}
\toprule
Predictor & Adjustment & AUC & LOSS\\
\midrule
COTTI w/o ABA & None & 0.8102 & 0.1496\\
COTTI w/o ABA & Time only & 0.8108 & 0.1487\\
COTTI w/o ABA & Affine & 0.8102 & 0.1489\\
COTTI w/o ABA & Uncertainty only & 0.8116 & 0.1485\\
TEN & ABA & 0.8123 & 0.1486\\
COTTI (Full) & ABA & \textbf{0.8127} & \textbf{0.1482}\\
\bottomrule
\end{tabular}
\end{table}

ABA improves the COTTI predictor's AUC by 0.0025 and reduces loss by 0.0014 relative to no adjustment. The strongest simpler control, uncertainty-only adjustment, reaches 0.8116 AUC and 0.1485 loss; adding event dependence yields a further 0.0011 AUC gain and 0.0003 loss reduction. Affine calibration reduces loss to 0.1489 while leaving AUC unchanged, consistent with a positive monotone mapping. The time-only AUC gain of 0.0006 can arise from changed ordering across times; a common bias still preserves candidate ordering within a request, as shown in Section~\ref{sec:properties}.

ABA also improves TEN from 0.8101 to 0.8123 AUC and from 0.1503 to 0.1486 loss. COTTI's remaining advantage over TEN with ABA is 0.0004 in both AUC and loss. Thus, event-conditioned adjustment benefits both predictors and substantially narrows the gap between them. The full model leads the tested controls, while the shared benefit of ABA limits attribution of the unadjusted comparator gap to CPC and RW alone. These system-level contrasts do not decompose additive component effects.

\subsection{Online Evaluation}

\subsubsection{Controlled A/B Experiment}
The A/B experiment ran from December 1 to 10, 2025, using 25\% of real-time traffic. HDR was the serving control and COTTI the treatment. We do not report the arm allocation, randomization unit, per-arm counts, or concurrent policy changes; the experiment description therefore does not support an independent assessment of balance and interference between arms.

The weighted-average improvements are +3.14\% CVR and +2.68\% RPM. The aggregation weights are not specified. These point estimates characterize the tested replacement of HDR by COTTI under the statistical scope described in the experimental setup.

\subsubsection{Choice of Online Control}
HDR is the serving policy replaced by the treatment, so this comparison evaluates the deployment decision. TEN and TEN with ABA are offline controls only; the online business-lift conclusion is restricted to HDR.

\subsubsection{Full Deployment}
\begin{figure}[t]
\centering
\includegraphics[width=\linewidth]{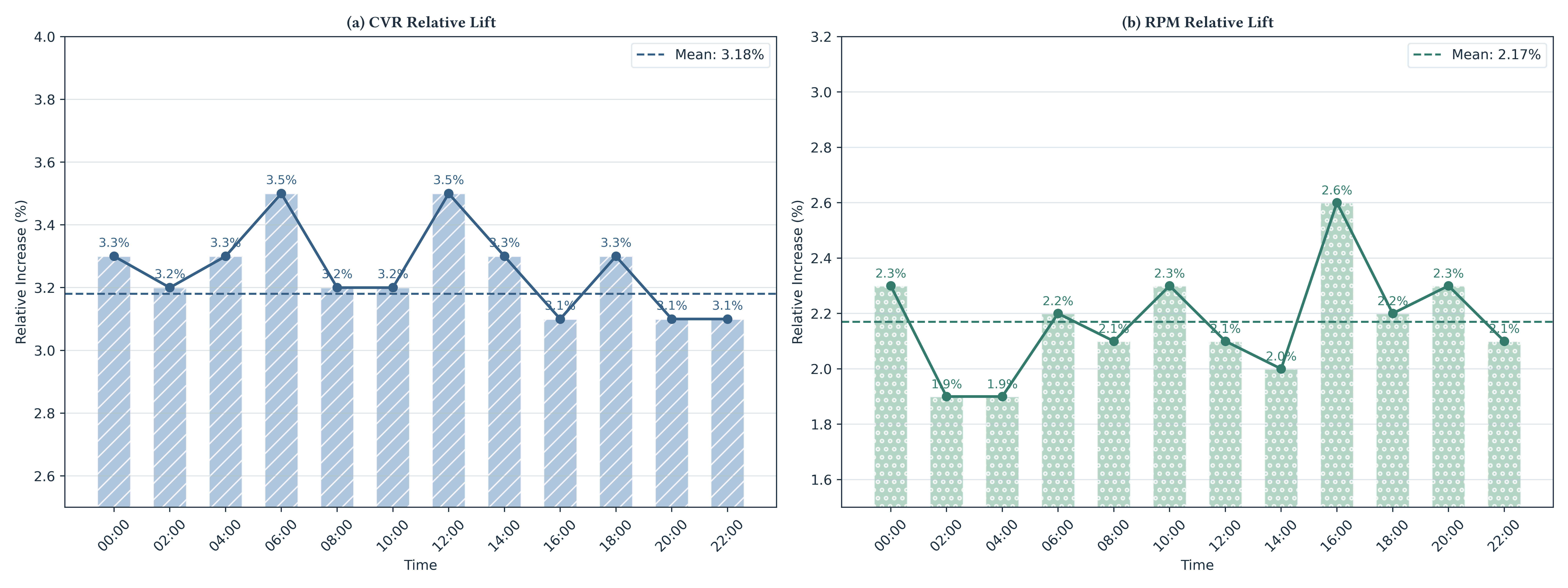}
\caption{CVR and RPM lift at two-hour intervals during December 12 full deployment. The trace is descriptive; its reference population and aggregation rule are unspecified.}
\label{fig:lift}
\end{figure}

COTTI was subsequently deployed to all real-time traffic on December 12, 2025. Figure~\ref{fig:lift} shows CVR and RPM lift at two-hour intervals. We treat this trace as descriptive because its reference population and aggregation rule are not specified.

Full rollout is distinct from the ten-day A/B experiment and supplies no concurrent control. Earlier-period comparisons may reflect traffic, inventory, prices, or campaign intensity. This trace is therefore descriptive, not an additional controlled estimate of treatment effect.

\section{Deployment and Efficiency}
\label{sec:deployment}

\subsubsection{Training and Serving Pipeline}
Training uses Alibaba MaxCompute logs and PAI in an environment with 20 parameter servers, 20 workers, and 64 NVIDIA A100 GPUs. Unspecified per-run allocations and stopping rules prevent resource-normalized cost comparisons. Data dependencies and configurations affect maintenance \cite{sculley2015debt}.

Serving uses Alibaba's Real-Time Prediction infrastructure with feature-store user profiles, candidate itineraries, and historical features. Hardware, instances, load, and candidate scale are matched; hardware specifications, precision, and batching remain unspecified. CPC and RW affect training; ABA adjusts scores without request-time parameter updates.

\begin{table}[t]
\centering
\small
\caption{Training time and single-value inference latency. The latency aggregation statistic and timing boundary are unspecified; no percentile interpretation is assigned.}
\label{tab:reported_efficiency}
\begin{tabular}{lcc}
\toprule
Method & Training time & Inference latency\\
\midrule
Base & 78m58s & 28.82\,ms\\
HDR & 80m42s & 29.17\,ms\\
ES-DFM & 78m03s & 28.74\,ms\\
FNW & 85m16s & 29.16\,ms\\
TEN & 81m22s & 26.31\,ms\\
COTTI & 83m17s & 28.62\,ms\\
\bottomrule
\end{tabular}
\end{table}

Table~\ref{tab:reported_efficiency} reports COTTI training/inference at 83m17s/28.62\,ms, versus Base's 78m58s/28.82\,ms. TEN has the lowest latency (26.31\,ms). Table~\ref{tab:efficiency_eval} specifies the end-to-end timing boundary for complete stochastic-serving cost.

\subsubsection{Stochastic Prediction Path}
The estimator shares one deterministic encoder output across $M=10$ head samples with dropout rate 0.2. One deterministic head pass produces $h_{ui}$; $M$ stochastic passes estimate uncertainty. Computation is approximately $C_{\mathrm{enc}}+(M+1)C_{\mathrm{head}}$, versus $C_{\mathrm{enc}}+C_{\mathrm{head}}$ without ABA. This excludes feature retrieval and scheduling overhead and does not determine elapsed latency.

Parallelism can reduce elapsed time without reducing computation. Table~\ref{tab:efficiency_eval} includes feature retrieval, the deterministic base prediction, and all ten stochastic ABA predictions. Its end-to-end measurements differ from Table~\ref{tab:reported_efficiency}'s single-value latency statistic.

\begin{table}[t]
\centering
\small
\caption{Measured end-to-end serving latency (ms) and throughput (requests/s). Hardware, instance count, load, and candidate scale are matched; QPS uses a common latency constraint.}
\label{tab:efficiency_eval}
\setlength{\tabcolsep}{3pt}
\begin{tabular}{lcccc}
\toprule
Configuration & p50 & p95 & p99 & QPS\\
\midrule
Base & 26.6 & 33.7 & 42.1 & 14,500\\
HDR & 27.0 & 34.2 & 42.8 & 14,000\\
TEN & 24.3 & 30.8 & 38.2 & 15,800\\
COTTI, ABA off & 24.8 & 30.3 & 36.7 & 15,400\\
COTTI, $M=10$ & 26.4 & 32.0 & 38.9 & 14,700\\
\bottomrule
\end{tabular}
\end{table}

\subsubsection{Load and Cost Evaluation}
At matched load and a common throughput latency constraint (Table~\ref{tab:efficiency_eval}), ABA adds 1.6/1.7/2.2\,ms to p50/p95/p99 and reduces throughput from 15,400 to 14,700 requests/s (4.5\%). Full COTTI has lower p99 than Base (38.9 versus 42.1\,ms) and HDR (42.8\,ms), and higher throughput than both. Compared with TEN, p99 is 0.7\,ms higher and throughput 1,100 requests/s (7.0\%) lower. ABA thus incurs measurable cost; efficiency is not uniformly superior.

The latency threshold, concurrency, warmup, duration, and repetitions remain unspecified; this single $M=10$ operating point establishes neither optimal sampling nor resource utilization.

\section{Conclusion}

COTTI combines cross-period alignment, historical-distance reweighting, and event-conditioned scoring. In the tested setting, aggregate predictive and business gains coexist with lower novel-category recall and higher serving cost. ABA also benefits TEN, narrowing COTTI's advantage. These trade-offs motivate jointly assessing aggregate quality, discovery, and deployment efficiency.\label{sec:body_end}

\bibliographystyle{IEEEtran}
\bibliography{references}
\end{document}